\documentclass{article}

\usepackage[nonatbib, preprint]{neurips_2024}

\usepackage[utf8]{inputenc} 
\usepackage[T1]{fontenc}    
\usepackage{hyperref}       
\usepackage{url}            
\usepackage{booktabs}       
\usepackage{amsfonts}       
\usepackage{nicefrac}       
\usepackage{microtype}      
\usepackage{xcolor}         
\usepackage{graphicx} 
\usepackage{algorithm}
\usepackage{algorithmic}
\usepackage{amsmath} 
\usepackage{multirow} 
\usepackage{caption}
\usepackage{xcolor}
\usepackage{ulem}    
\usepackage{tabularx}

\usepackage{fancyhdr}
\usepackage{datetime2} 
\newcommand{\ra}[1]{\renewcommand{\arraystretch}{#1}}

\title{GenMM: Geometrically and Temporally Consistent Multimodal Data Generation for Video and LiDAR } 
\author{%
  Bharat Singh\thanks{Equal contribution}  \\
  \And 
  Viveka Kulharia$^*$ \\
  \And
  Luyu Yang$^*$ \\ 
  \AND
 Avinash Ravichandran \qquad Ambrish Tyagi \qquad Ashish Shrivastava \vspace{0.1in} \\ 
  Cruise LLC \vspace{0.05in} \\
 \texttt { \{bharat.singh, viveka.kulharia, luyu.yang, avinash.ravichandran, } \\ 
 \texttt {ambrish.tyagi, ashish.shrivastava\}@getcruise.com}
}
  
\newcommand{\lidar}{LiDAR}

\begin{document}

\maketitle
\begin{abstract}
  Multimodal synthetic data generation is crucial in domains such as autonomous driving, robotics, augmented/virtual reality, and retail. We propose a novel approach, GenMM, for jointly editing RGB videos and \lidar{} scans by inserting temporally and geometrically consistent 3D objects. Our method uses a reference image and 3D bounding boxes to seamlessly insert and blend new objects into target videos. We inpaint the 2D Regions of Interest (consistent with 3D boxes) using a diffusion-based video inpainting model. We then compute semantic boundaries of the object and estimate it's surface depth using state-of-the-art semantic segmentation and monocular depth estimation techniques. Subsequently, we employ a geometry-based optimization algorithm to recover the 3D shape of the object's surface, ensuring it fits precisely within the 3D bounding box. Finally, \lidar{} rays intersecting with the new object surface are updated to reflect consistent depths with its geometry. Our experiments demonstrate the effectiveness of GenMM in inserting various 3D objects across video and LiDAR modalities.

\end{abstract}

\section{Introduction}
Cameras and \lidar{} are two pivotal optical sensing modalities widely used in applications such as autonomous vehicles (AV), augmented/virtual reality (AR/VR), and robotics~\cite{hu2023_uniad, Roberts_2021_ICCV, sun2020scalability}. These modalities complement each other: \lidar{} provides accurate depth information, while cameras offer rich semantic details~\cite{transfusion_2022, Chen2016Multiview3O, point_painting_2020,yin2024isfusion} for improved visual understanding.
As machine learning models that process these modalities grow in size, complexity, and performance requirements, the need for synthetic data has become increasingly evident~\cite{carla_2017,virtual_kitti,simgan_2017_CVPR}.
This is crucial to address the challenge of lack of data for scenarios with limited data diversity or long-tail parts of the data distribution.

In applications like AVs and AR/VR, generating content for both image and \lidar{} modalities is essential to ensure realistic object placement and perception~\cite{arkitscenes}. Synthetic data generation offers significant potential for enhancing existing video and \lidar{} datasets by increasing object diversity. For instance, existing objects (e.g., cars, pedestrians) in datasets can be replaced with new ones having different appearances or poses. This capability is particularly valuable for creating new scenarios that aid in long-tail data generation, where data collection is challenging or risky. By inserting new objects into existing video and \lidar{} datasets, we can simulate rare or hazardous situations, thus improving the robustness of machine learning models.

While simulation~\cite{Manivasagam2020LiDARsimRL} and neural rendering (e.g., ~\cite{yang2023unisim}) are popular methods for inserting objects into scenes, they often require rendering the entire scene, which leads to a loss of background information, especially in high-resolution scenarios. This increases the distribution gap between real and generated samples. Additionally, neural rendering approaches rely on multiple views of an object and struggle with relighting new objects, generating varied motion patterns for articulated objects (e.g., people, animals) or temporally varying albedo (like flashing lights).

Recent generative models such as ~\cite{ho2020denoising, rombach2022high} can perform realistic local edits to scenes while retaining high background fidelity. However, these models focus on generating images only, are not geometrically grounded and result in temporally inconsistent outputs for object insertion or replacement. Furthermore, rendering articulated objects or generating accurate object shapes with these models is still challenging. To this end, methods such as ~\cite{geyer2023tokenflow, chen2023control, controlnet_iccv_2023} have explored the use of image latents/control primitives like pose, edges, and depth maps, however, obtaining these surrogates can be difficult when inserting a wide range of novel objects such as humans, vehicles, furniture, {\it etc} in new scenes.

\begin{figure}[h]
\centering{
\includegraphics[width=\linewidth]{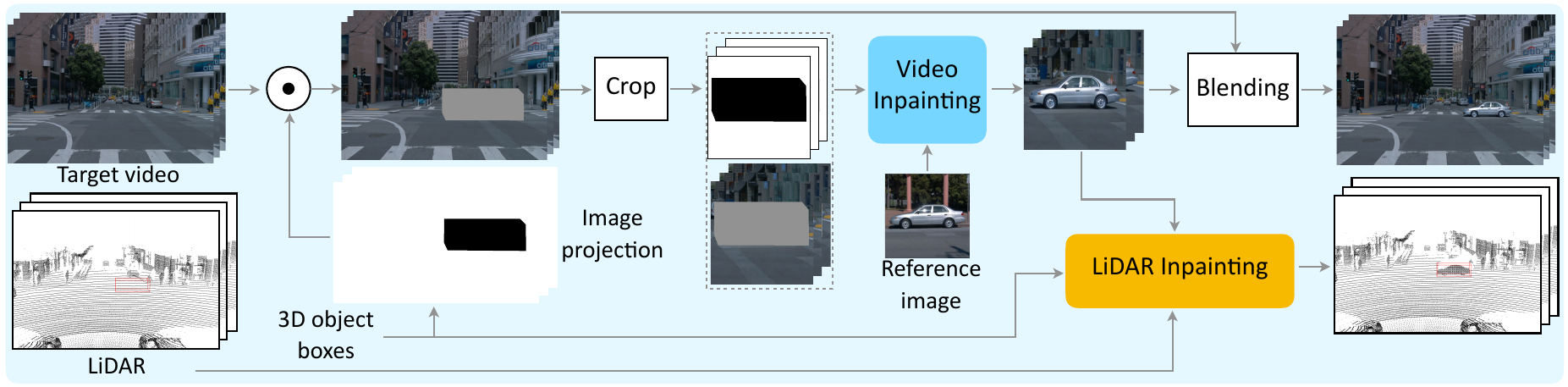}}
\caption{Overview of the proposed GenMM method. Given a target video and its corresponding \lidar{} frames, we  project 3D boxes onto the 2D image to create masked RoIs, which are combined with the original image to produce masked input images. These cropped masked inputs, along with their respective masks, are then processed through our video inpainting method (Figure~\ref{fig::video_inpaint}), which generates frames with the inpainted objects. These inpainted crops are then blended back into the target video. The inpainted crops and the \lidar{} point cloud are input to a geometry-based \lidar{} inpainting algorithm (Figure~\ref{fig::lidar_inpaint}) that generates the corresponding \lidar{} points for the inserted objects.
}
\label{fig::pipeline}
\end{figure}

In this paper, we introduce GenMM, a method for simultaneously editing  RGB videos and \lidar{} scans by incorporating 3D objects that maintain both temporal and geometric consistency. Figure~\ref{fig::pipeline} provides an overview of our method. GenMM uses a reference image and a 3D bounding box sequence to seamlessly integrate objects into video sequences. By projecting the 3D bounding boxes corresponding to the object's track in the video, we identify Regions of Interest (RoIs) and employ a diffusion-based video inpainting model to accurately inpaint these regions (Sect.~\ref{subsec:method_video_inpainting}). We then utilize state-of-the-art semantic segmentation and monocular depth estimation methods to compute the semantic boundaries and surface depth of the inserted objects. Next, to generate \lidar{} data corresponding to the inserted object, we use a geometry-based optimization algorithm (Sect.~\ref{subsec:method_lidar_inpainting}) to align the 3D surface of the generated object by ensuring they fit  within the specified 3D bounding box. We update the \lidar{} rays intersecting with the new object surface to ensure depth consistency with the objects' geometry.

To the best of our knowledge, this is the first work using a diffusion model to generate both video and its corresponding \lidar{} data.
Our contributions are summarized below:
\begin{itemize}
    \item We propose a novel method for generating multimodal data, which includes both image and \lidar{} modalities by leveraging recent progress made in diffusion and foundation models.
    \item We develop a novel video inpainting algorithm for coherently inserting an object within a video using its reference image.
    \item We create a geometry-based \lidar{} inpainting algorithm that produces \lidar{} outputs corresponding to the generated video but operates independently of the video inpainting algorithm. This allows us to leverage the latest advances in video inpainting methods.
\end{itemize}

\section{Related Work}
Graphics-based simulation engines such as Unity3D\cite{unity3d}, Unreal Engine~\cite{unrealengine}, and Omniverse~\cite{omniverse, savva2019habitat} are popular choices for generating multimodal synthetic data for many applications like robotics or AR/VR. However, they struggle with sensor fidelity, primarily because rendering scenes requires reconstructing geometric and material properties, recovering light sources and finally performing accurate ray-tracing. 
 Additionally, these engines render entire scenes, leading to unnecessary loss of background information when only a few objects need to be added.
LiDARSim~\cite{Manivasagam2020LiDARsimRL} can generate entire scenes using 3D assets. 
However, since it relies on ray casting within the graphics engine, the fidelity of corresponding videos is constrained by the capabilities of the graphics engine. Further, this method needs 3D assets which can be time consuming and expensive to create. 

Neural rendering~\cite{mildenhall2021nerf, wu2023mars, yang2023emernerf,yang2023unisim} and Gaussian Splatting~\cite{zhou2023drivinggaussian} are also popular for integrating objects into scenes, but they suffer from computational inefficiency, particularly in generating high-resolution images. 
As these methods do not model light sources, they have fidelity issues like poor shadow rendering when objects are added or when the lighting does not align between inserted objects and the background. The dependence of only training on a single video and lack of a world model of how objects move also complicates their use with articulated objects (e.g., pedestrians, animals) or in dynamic scenes (e.g., vehicles with turn signals or flashing lights). There have also been efforts to insert 3D objects in existing scenes. For example, \cite{chen2021geosim} places 3D assets in the scene and then uses image synthesis networks for blending. Lift3D/GINA-3D \cite{li2023lift3d,shen2023gina} on the other hand utilize GANs and NeRFs to place objects in 3D scenes. However, these methods are limited to scenarios where we can build a multi-view library of assets and have similar drawbacks as neural rendering techniques.

Recently, video-based diffusion models have emerged as open-world simulators capable of synthesizing photo realistic data in high resolution~\cite{blattmann2023stable,videoworldsimulators2024, guo2023animatediff}.
Prior art, such as text-based scene generation~\cite{hu2023gaia}, often lacks practicality for tasks like inserting or swapping objects within a scene. 
Although there are methods for animating static images into videos~\cite{guo2023animatediff, hu2023animate}, the specific challenge of object insertion in video contexts which adheres to geometrical properties of the scene remains relatively unexplored.

Methods such as ~\cite{poole2022dreamfusion,tang2023make,wang2023score} have also used diffusion models to estimate the 3D geometry of objects.
However, they primarily focus on individual objects and the synthesis of novel views. 
In contrast, our work aims to construct complete 3D scenes by inserting an object into the specified scene.

\section{Method}
Our approach requires the following inputs: a reference image of the object to be inserted, 3D bounding boxes, target video frames, and \lidar{} point clouds. The goal is to place the reference object within the scene to produce a temporally consistent video and a geometrically consistent \lidar{} output, ensuring that the generated data closely matches the appearance of the reference image. Figure~\ref{fig::pipeline} illustrates the various steps of the proposed approach.

Given a sequence of 3D bounding boxes, represented by their 3D position ($x_t,y_t,z_t$), size ($l,w,h$), and orientation ($\psi_t$, $\phi_t$), our method first obtains 
their Regions of Interest (RoIs) by projecting these boxes onto camera frames. These RoIs are where objects are inserted in the video using a diffusion-based video Inpainting-Unet. 
Appearance features (i.e CLIP features \cite{RadfordCLIP}) computed from the reference image of the object are used for spatial and temporal attention to ensure that the appearance of the inserted object in the video closely resembles the reference image.
After insertion, we obtain the object's semantic boundary and remove non-object voxels~\cite{kutulakos2000theory} from the 3D bounding box. 
 We estimate the inserted object’s relative depth~\cite{yang2024depth} and then convert it to metric depth by using the \lidar{} points in the background to get the scale and shift parameters. We further refine the object surface to ensure that it fits inside the 3D bounding box using a constrained optimization. Finally, the \lidar{} rays that intersect with the object's surface (represented as 3D voxels) are updated to reflect the geometry of the inserted object. 
The subsequent sections provide more details.
\begin{figure}[b]
\centering{
\includegraphics[width=0.95\linewidth]{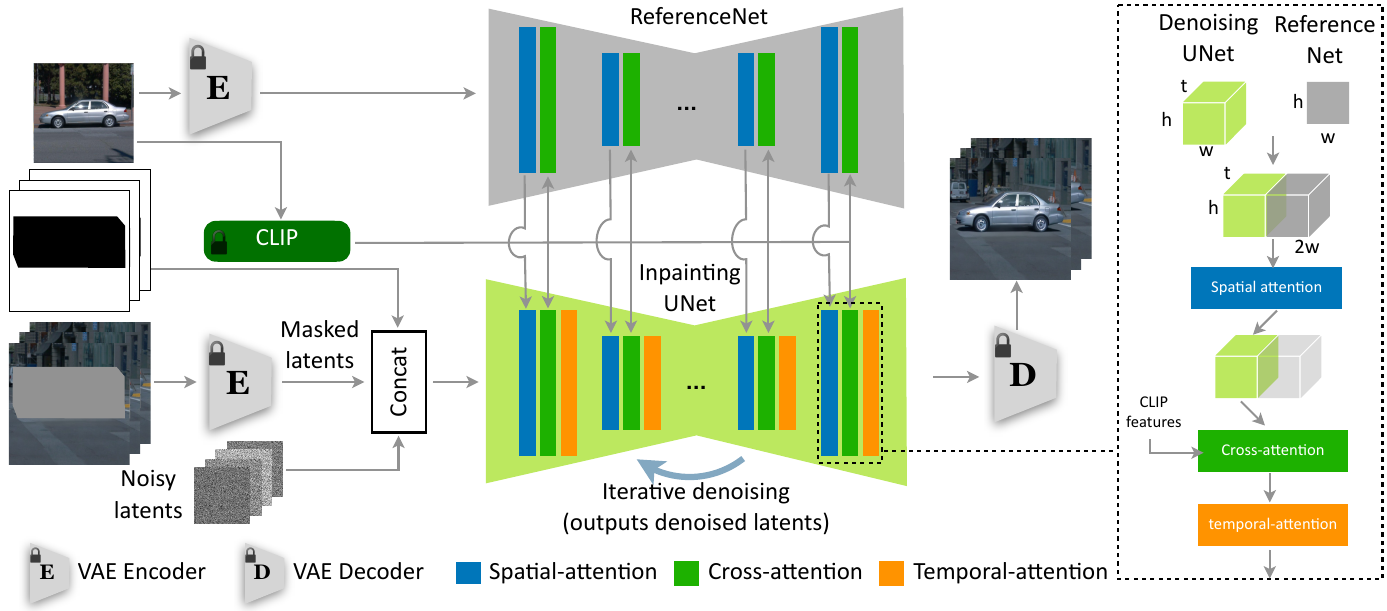}}
\caption{Video inpainting using a reference image. 
We inpaint objects using a reference image, object masks, and a masked input image to ensure realistic object insertion. 
The Inpainting-Unet network utilizes concatenated features from the object mask, masked latents, and noisy latents. 
Following the approach in ~\cite{hu2023animate}, we employ spatial-attention layers (using ReferenceNet features) to ensure appearance consistency between the reference image and the inpainted objects, along with temporal-attention layers to maintain temporal consistency.
}
\label{fig::video_inpaint}
\end{figure}

\subsection{3D Object Placement and Generating Mask Sequences}
We start by placing a sequence of 3D bounding boxes in the scene. These bounding boxes can be obtained using manual annotations or automatically using scenario generation algorithms~\cite{chen2021geosim,suo2021trafficsim,schulz2018interaction}. Our algorithm is agnostic to the source. We insert an object into these locations in a given sequence using a reference image. When selecting the object reference image, the pose of the object should be similar to the yaw ($\psi$) of the 3D box sequence. 

We project the sequence of 3D bounding boxes onto the image. This projection results in $8$ points corresponding to each corner of the 3D box in the image. If we were to create a 2D bounding box that encloses these $8$ points, it would be much larger than the actual boundary of the object in the perspective view. Hence, to avoid inpainting a much larger region than necessary, we use a non-rectangular RoI mask that corresponds to the perspective projection of the 3D bounding box.

\subsection{Video Inpainting}
Our method for video inpainting builds on the AnimateAnyone architecture~\cite{hu2023animate}, which is a derivative of AnimateDiff~\cite{guo2023animatediff}. We keep the ReferenceNet and CLIP based conditioning from~\cite{hu2023animate} and propose to use an Inpainting-Unet for video synthesis as shown in Figure~\ref{fig::video_inpaint}. We drop the Pose Guider network from \cite{hu2023animate} as this information is not available for many objects and we also want to remove control primitives. Unlike CLIP, ReferenceNet computes high-resolution features of the reference object at various scales, and by conditioning on it, the Inpainting-Unet learns to fill in pixels in target frames.

For the Inpaining-UNet, the input is created by concatenating the latent VAE features of the background image (the foreground pixels are greyed out) with the inpainting mask and latent noise, as done in~\cite{rombach2022high}. The first layer is different in this Inpainting-Unet compared to a text-to-image diffusion model as the input channels are different (9 instead of 4), otherwise, the remaining network is identical. Our contributions lie in adapting~\cite{hu2023animate} to create a generic video Inpainting-Unet while simplifying the control conditions of the original method, which was originally employed for the specific task of pose-guided video synthesis.

\subsubsection{Preparing Training Data for Video Inpainting}
The training data for the video Inpainting-Unet consists of videos with object tracks defined by 2D bounding boxes. This allows us to train our Inpainting-Unet on any video, rather than limiting us to datasets with 3D annotations. We only use the 3D boxes and their masks during inference.

We create binary masks for each frame using 2D bounding boxes and enlarge them by 10\% to ensure sufficient context around the object for rendering shadows. Finally, we crop a square region which encloses this added context and resize it to $512\times512$ pixels while training, like \cite{singh2018sniper}.

We insert objects into a target region, so we only need to learn to blend them with the surrounding context, not generate the entire frame. Therefore, we provide the appropriate context needed for video inpainting and do not train on full-resolution videos. Training on cropped regions instead of full-resolution images also reduces computational and memory costs. Additionally, it preserves the detail of the original content in the target frame outside the inpainted area.

It is also important to crop the region around the 2D object mask in each frame instead of using a predefined spatial RoI for the entire set of frames. This cropping strategy helps align the features for the 1D temporal transformer within the temporal-attention layers. With stronger backbones that perform attention across all dimensions ($x, y, t$), this may not be necessary, but for computationally efficient architectures, this is an effective strategy.

\subsubsection{Two Stage Training of the Video Inpainting-Unet}
\label{subsec:method_video_inpainting}
The training of our video Inpainting-Unet (Figure~\ref{fig::video_inpaint}) consists of two stages. In the first stage, we omit training the temporal-attention modules and focus on learning to replicate the appearance of the reference image against a different background and within a 2D bounding box of varying size and shape. To facilitate this training, we sample two crops from the same object track in a video, ensuring they are less than 10 frames apart. One of these sampled crops serves as the reference image, while the other functions as the target output.
Following the standard practice in diffusion-based inpainting (e.g., ~\cite{rombach2022high}), we mask the target image at the 2D bounding box location by graying out the masked pixels (i.e., setting the pixel values to $128$). 
The inpainting network takes the mask, the masked target image, and a noised version of the target image as inputs, concatenating them across the channels in the latent space of a VAE encoder. 
Finally, we optimize the following loss to update the Inpainting-Unet and the ReferenceNet model parameters:
\begin{equation}
    \min_{\boldsymbol \theta, \boldsymbol \phi} \|\epsilon - \epsilon_{\boldsymbol \theta}(\boldsymbol z_t, g_{\boldsymbol \phi}(\boldsymbol x_{\text{ref}}), c(\boldsymbol x_{\text{ref}}), t)\|_2^2,
\end{equation}
where $\boldsymbol \theta$ are the Inpainting-Unet parameters, $\boldsymbol \phi$ are the ReferenceNet parameters, $\boldsymbol z_t$ denotes the noisy latent after $t$ diffusion steps, $g_{\boldsymbol \phi}(\boldsymbol x_{\text{ref}})$ represents the ReferenceNet features for spatial-attention, and $c(\boldsymbol x_{\text{ref}})$ represents the CLIP features for cross-attention.

In the second stage, our goal is to enhance the temporal consistency of generated objects. 
Hence, we augment the Inpainting-Unet model by incorporating temporal-attention layers  (\cite{guo2023animatediff, hu2023animate}). 
We randomly select object tracks consisting of $10$ frames each, using the first frame as a reference frame, and then train only the temporal layers of the Inpainting-Unet to generate all $10$ frames while keeping other parameters (including $\boldsymbol \theta$ and $\boldsymbol \phi$) frozen.
We tried to combine the two stages in a single training, however, we find that training them sequentially leads to better quality results, as also shown in \cite{hu2023animate}.

\subsection{\lidar{} Inpainting Using the Generated Video}
\label{subsec:method_lidar_inpainting}

\begin{figure}[t]
\centering{
\includegraphics[width=\linewidth]{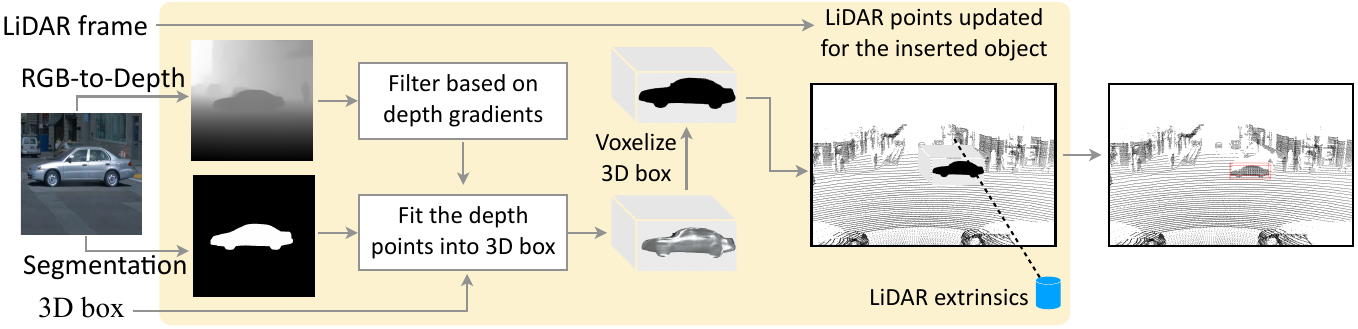}}
\caption{Overview of geometry-based LiDAR inpainting approach. Pixels on inserted 2D object are lifted to 3D with proper scale and shift of depth to fit the target 3D bounding box. 
We voxelize the 3D bounding boxes to represent the 3D object surface. 
Each voxel within the bounding box is classified as either occupied or empty. \lidar{} rays that intersects with an occupied voxel are updated with the correct range corresponding to the inserted object.
}
\label{fig::lidar_inpaint}
\end{figure} 

To insert the reference object in the \lidar{} data (Figure \ref{fig::lidar_inpaint}), we first estimate the object surface and then update the range of those \lidar{} rays which intersect with this surface. To estimate the object surface, we utilize the DepthAnything model~\cite{yang2024depth} to generate the relative depth of each pixel within the corresponding 2D inpainted image region. Due to the inherent scale ambiguity in monocular RGB-to-depth models, we determine a global scale for the object to ensure that the points corresponding to the object are fully contained within the 3D bounding box.

\textbf{Filtering object points}: To identify the points belonging to the inserted object, we utilize Grounding Dino~\cite{liu2023grounding} to get a tight crop around the object, followed by the Segment Anything Model (SAM)~\cite{kirillov2023segment} to obtain the semantic boundaries of the object. While SAM performs well in segmenting object pixels, some boundary pixels may still end up in the background region, potentially leading to ``leaking depth'' artifacts. To mitigate this artifact, we calculate the depth gradients of the depth image and filter out points with large depth gradients (refer Figure \ref{fig::lidar_inpaint}).

\textbf{Estimating metric depth}:
Let $d_i$, $i=1, \dots, N$, denote the relative depths of the filtered pixels of a segmented object. These relative depths are inverted disparity values obtained from DepthAnything. To recover metric depth, we apply a global affine transformation on the relative depth, whose parameters, scale $\alpha$ and shift $\beta$, are estimated using the background \lidar{} points in the target frame using RANSAC~\cite{ransac_1981}. We further refine these parameters so that the metric depth values accurately fit within the specified 3D bounding box using the following optimization problem:
\begin{equation}
    \max_{\alpha, \beta} \alpha \quad \text{subject to} \;\; \delta_{min} \le X_i (d_i \alpha + \beta) \le \delta_{max}, \forall i \in N \quad \text{and} \;\; \alpha > 0,
\end{equation}
where $X_i$ represents the 3D position of the $i^{th}$ pixel, which is computed by first lifting the pixel coordinates to 3D space by multiplying it with the inverse of camera intrinsic matrix, $\mathbf{K}^{-1} \in \mathbb R^{3 \times 3}$, and then applying a rotation matrix, $\mathbf{R} \in \mathbb R^{3 \times 3}$, to align the points with the coordinates of the 3D bounding box. 
The transformation can be expressed as: $X_i = \mathbf R* \mathbf K^{-1} * [x_i, y_i, 1]^T$,
where $x_i \in [0, W-1]$ and $y_i \in [0, H-1]$ denote the pixel locations of the $i^{th}$ point in the image, and `$*$' denotes  matrix multiplication.
Here $W$ and $H$ are the width and height of the camera image, respectively.
After scaling, all depth points should be  within the 3D bounding box. 
This constraint is captured by the expressions $\delta_{\text{min}} = \mathbf{R} * \boldsymbol{b} - \boldsymbol{b}_0$ and $\delta_{\text{max}} = \mathbf{R} * \boldsymbol{b} + \boldsymbol{b}_0$. 
Here, $\boldsymbol{b}$ represents the location of the 3D bounding box center in camera coordinates, and $\boldsymbol{b}_0$ is defined as $[l/2, w/2, h/2]$, where $l$, $w$, and $h$ are the length, width, and height of the 3D bounding box, respectively.

\textbf{Updating \lidar{} Range}:
We start by voxelizing the 3D bounding box of the inserted object, assigning a value of $1$ to voxels that contain 3D points on the object's surface and setting others to $0$. Subsequently, for each \lidar{} beam, we update its range if it intersects with an occupied voxel (corresponding to our object). In scenarios where the beam intersects multiple occupied voxels, the range is updated to be the closest occupied voxel distance to the \lidar{} sensor.

\section{Experiments}
We train our video inpainting diffusion models using the BDD100K~\cite{yu2020bdd100k} and the Waymo Open~\cite{sun2020scalability} Datasets. 
As 3D information about the scene and cameras is available only in the Waymo Open Dataset, we conduct our multi-modal evaluations exclusively on scenes from this dataset. 

For the first stage of training, we select 2D tracks of various objects from video sequences, excluding those that are very small. We sample pairs of frames from a tracklet that are within 10 frames of each other, yielding 8,960 tracklets from the BDD dataset and 9,344 from the Waymo dataset. In the second training stage, we choose an arbitrary frame from a tracklet as the reference frame and mask out the object in the next 10 frames. This stage requires tracklets to be of a minimum length of 10 frames, resulting in 2,080 tracklets from the BDD dataset and 4,256 from the Waymo dataset. We use an 80/20 split for training and testing in both stages.

We train the first stage for 80,000 iterations using a batch size of 16 images per GPU, and the second stage for 40,000 iterations using a batch size of 4 images per GPU on four H100 GPUs.
We initialize our Inpainting-UNet model using the RealisticVision-5.0-inpainting model and the ReferenceNet with the RealisticVision-5.0 model which are available on HuggingFace. Other hyper-parameters are same as the publicly available implementation. During inference, we require a target video, 2D ROIs (projections of 3D bounding boxes), and a reference image for the object to be inserted. 
 
\subsection{Video Inpainting}
We evaluate our inpainting method for tasks pertaining to animating, swapping, and inserting reference objects in video. We compare our video inpainting technique with LoRA~\cite{peft} based frame level inpainting and an inpainting version of AnimateDiff~\cite{guo2023animatediff} which is conditioned on CLIP features of a reference image. These methods are denoted as LoRA and AnimateDiff-CLIP in Table~\ref{fig::table}, respectively. For the LoRA baseline, we use Dreambooth~\cite{ruiz2023dreambooth} based LoRA training~\cite{hu2021lora} on the reference image for 1,200 iterations and then perform per-frame inpainting using this model. To evaluate the effectiveness of our video generation model, we compute the following three metrics: cross-frame Structural Similarity Index Measure (SSIM)~\cite{sara2019image}, Learned Perceptual Image Patch Similarity (LPIPS)~\cite{zhang2018perceptual} and Fréchet Video Distance (FVD)~\cite{unterthiner2019fvd}, the video version of the Fréchet Image Distance. To better measure the quality of video synthesis, we take the union of the object track which provides us with a union bounding box. We use this bounding box to crop the video and measure FVD on these  crops.

\begin{figure}[h]
\includegraphics[width=0.95\linewidth]{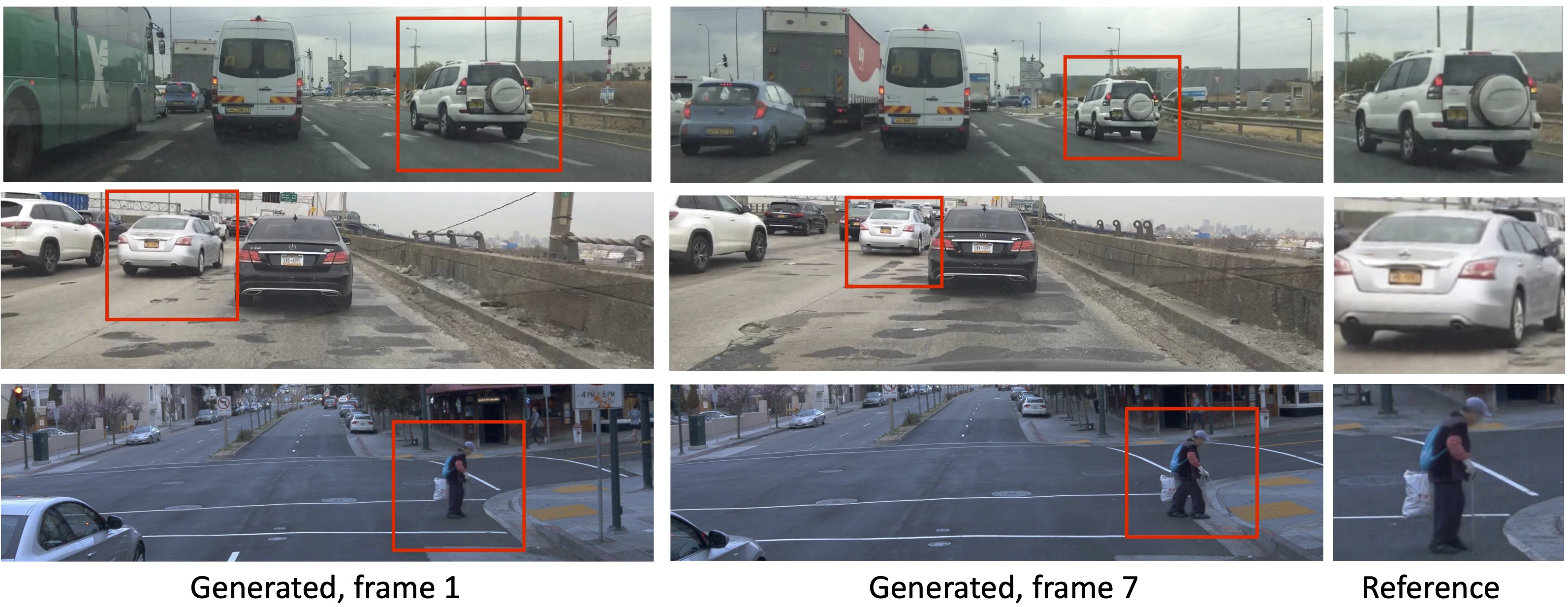}
\caption{Examples of animating reference crops in videos. Given a reference crop from first frame (right), we inpaint the object in subsequent frames. We mask out the RoI in the target image and generate the object inside the RoI conditioned on the reference crop. 
The model can learn to generate temporally consistent videos, without specifying control conditions like object pose or edge-map.
}
\label{fig::animate}
\end{figure}

\begin{table*}[b]
\centering
\setlength{\tabcolsep}{5pt}
\footnotesize
\ra{1.3}
\begin{tabular}{@{}lrrrrrrrrrr@{}}\toprule
& \multicolumn{3}{c}{\small{Animate first frame}} & \phantom{}& \multicolumn{2}{c}{\small{Swap objects in video}} &
\phantom{} & \multicolumn{2}{c}{\small{Insert objects in video}}\\
\cmidrule{2-4} \cmidrule{6-7} \cmidrule{9-10} & \small{LPIPS} $\downarrow$ & \footnotesize{SSIM} $\uparrow$ & \footnotesize{FVD} $\downarrow$ && \footnotesize{LPIPS-Ref} $\downarrow$ & \footnotesize{FVD}\footnotesize{ $\downarrow$ } && \footnotesize{LPIPS-Ref} $\downarrow$ & \footnotesize{FVD} $\downarrow$ \\ 
\midrule
\small{LoRA} & $0.31$ & $0.75$ & $939$ && $0.84$ & $1212$  && $0.82$ &  $1415$\\
\small{AnimateDiff-CLIP} & $0.23$ & $0.75$ & $499$ && $0.68$ & $556$ && $0.65$ &  $1232$\\
\small{GenMM (ours)} & $\textbf{0.10}$& $\textbf{0.89}$ & $\textbf{168}$ && $\textbf{0.61}$ & $\textbf{316}$ && $\textbf{0.54}$ & $\textbf{324}$\\
\bottomrule
\end{tabular}
\caption{Quantitative Results for animating, swapping, and inserting objects. Videos are evaluated for aesthetic quality using LPIPS, structural similarity using SSIM, and overall quality of video synthesis using FVD. 
For swapping and insertion, LPIPS is computed with reference image instead of the original video (denoted as LPIPS-Ref). SSIM is only computed for animate with ground truth.}
\label{fig::table}
\end{table*}

\paragraph{Animating Objects}
Figure~\ref{fig::animate} shows the ability of GenMM to take the first frame and then place the object in subsequent frames, also known as animating. As ground truth for a generic video synthesis task is hard to obtain, we choose this setting so that we can compare to ground truth. Here, we know the appearance of the object in the next few frames, which we use as ground truth. As shown in the figure, we can place cars in crowded and occluded scenes and also generate different poses of articulated objects like people.
We compute LPIPS and SSIM with respect to the original video frames and report them in Table~\ref{fig::table}. A lower LPIPS of $0.10$ and a higher SSIM metric of $0.89$, along with the qualitative results in Figure~\ref{fig::animate}, demonstrate that GenMM-synthesized videos for this task are almost indistinguishable from other real objects in the scene.

\begin{figure}[h]
\includegraphics[width=0.95\linewidth]{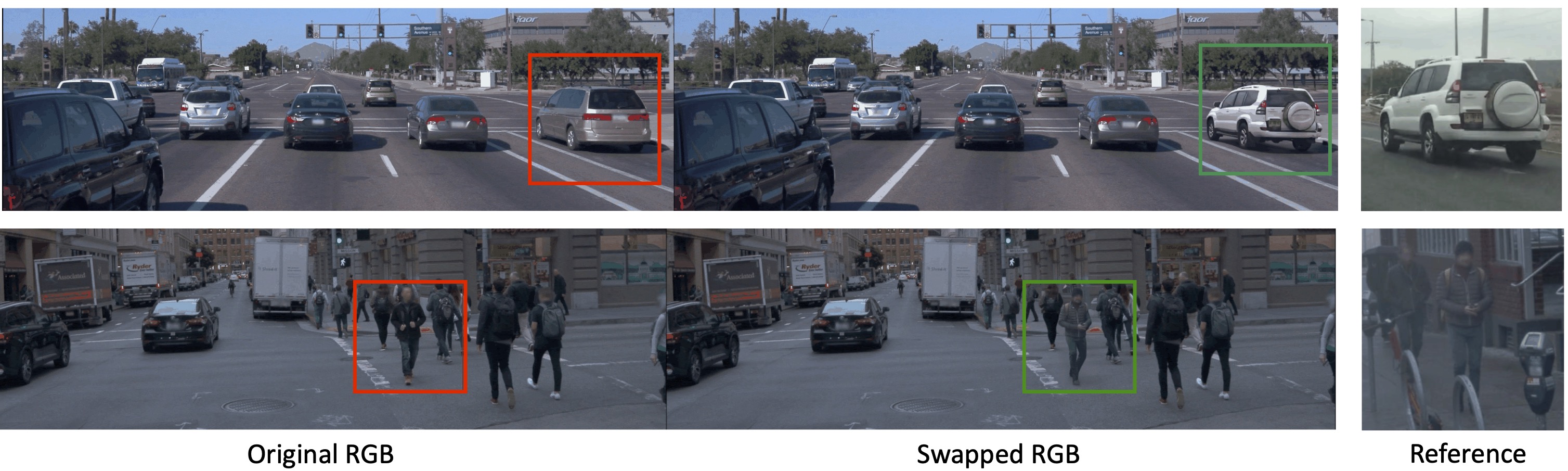}
\caption{Examples of swapping objects in videos. Each row shows a separate example. The red box around the object in the leftmost column image indicates the object that needs to be replaced, and the green box on the right images highlights the inserted object using reference. In the first row, we see that though the reference image is from a cloudy scene and does not have prominent shadows, our method is able to re-light the object in a new environment. We are also able to insert pedestrians and learn their walking patterns without specifying spatial controls, such as OpenPose or depth.}
\label{fig::swapping}
\vspace{-0.1in}
\end{figure}

\paragraph{Swapping Objects}
Fig \ref{fig::swapping} shows the ability of our method to take existing tracks of objects and replace them with other objects using their reference image alone. The results show that although the original object does not have shadows in the reference image, we are able to render photo-realistic shadows for the reference object, once it is inserted in a new scene. Similarly, for the second example, though the reference image has other people in the background, our method learns to focus on the center object and inpaint based on its appearance, even when background objects are present. Also, we do not require the source and target object to have the same shape, as is typically the case for methods which rely on structural conditions like edgemaps, depth, etc \cite{geyer2023tokenflow, controlnet_iccv_2023}. Quantitative results for object swapping is shown in Table \ref{fig::table}. In this case, we compute LPIPS with respect to the reference frame as we do not have ground truth for the object's appearance in the target video. This metric, which we call LPIPS-Ref, provides us an indication of the closeness of the generated crop to the reference frame (and is expected to be higher as we do not have the ground-truth video). We also measure FVD to evaluate the quality of video synthesis. We see that across metrics, our proposed approach outperforms other baselines, validating the efficacy of our method for swapping objects.

\begin{figure}[b]
\includegraphics[width=0.95\linewidth]{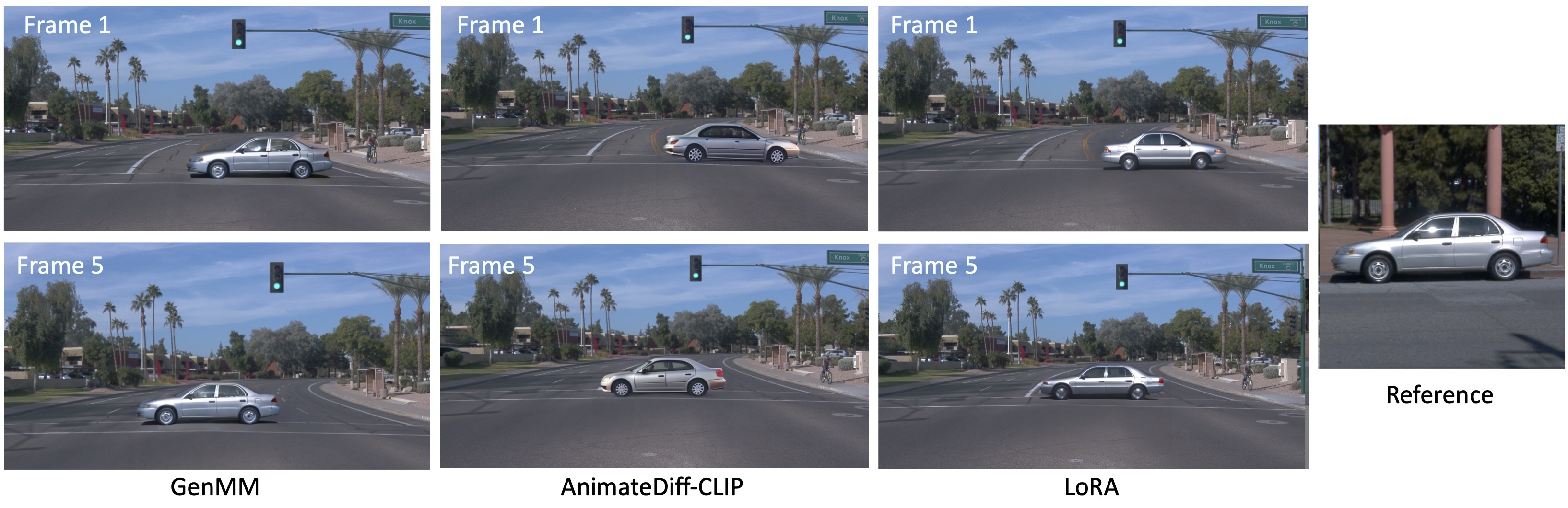}
\caption{Inserting objects: we show two different frames from a video sequence of GenMM against two other baselines: LoRA and an inpainting version of AnimateDiff~\cite{guo2023animatediff} trained on CLIP latents. The baseline methods can flip the pose while inpainting. AnimateDiff-CLIP~\cite{guo2023animatediff} also generates a different color from the reference image, as CLIP features do not capture that information accurately.}
\label{fig::comparison}
\end{figure}

\begin{figure}[t]
\includegraphics[width=0.95\linewidth]{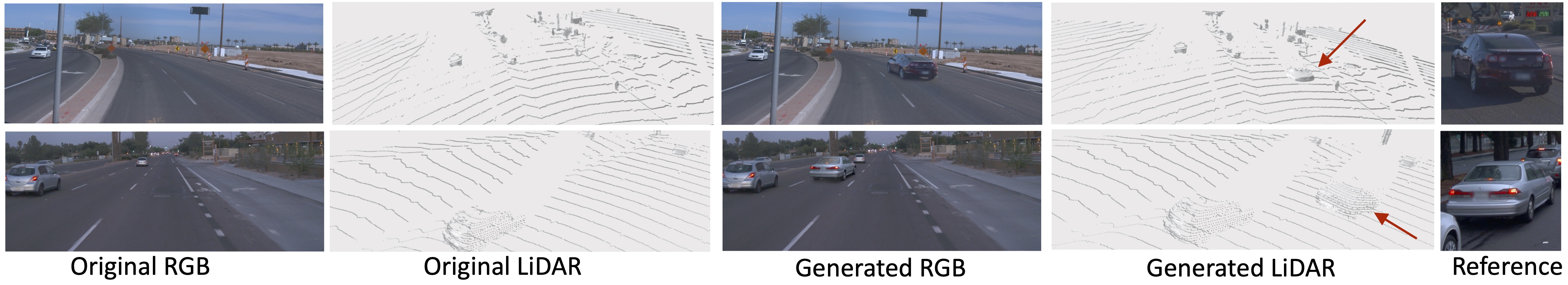}
\caption{Given a reference image (shown on the right), we are able to consistently place the object in the scene and generate its corresponding LiDAR points.}
\label{fig::lidar}
\end{figure} 

\paragraph{Inserting Objects}
Lastly, we show results for the insertion case, i.e., we take a reference image and insert it into a new scene with a novel 3D trajectory. We show the rendered objects after inserting in the video in Fig.~\ref{fig::lidar} and measure quantitative metrics in Table~\ref{fig::table}. We use the same metrics as we used in the swapping case. Fig~\ref{fig::comparison} shows another comparison of object insertion with GenMM with baseline methods. Since CLIP embeddings do not preserve the pose and color of the object, the appearance/pose of the object is sometimes flipped after inpainting. Further, the LoRA based baseline does not have a motion module, hence the size of the object is not consistent across frames, and this leads to a flickering pattern. However, LoRA better captures the appearance of the reference object compared to AnimateDiff-CLIP. GenMM leads to better aligned poses and temporally coherent results as it has access to high resolution features of the reference object from the ReferenceNet and has also learnt motion patterns of similar looking objects. Other methods like \cite{geyer2023tokenflow} do not work for object insertion due to a lack of structural conditioning or aligned latent features in the source video and end up generating content similar to the background video even when we used LoRA models trained on the reference object. We show an example for~\cite{geyer2023tokenflow} in our Appendix.

\begin{table*}
\centering
\setlength{\tabcolsep}{5pt}
\small
\ra{1.3}
\begin{tabular}{@{}lcccccc@{}}\toprule
& AbsRel error (a) $\downarrow$ & AbsRel error (b) $\downarrow$ &&   $l_2$ error (a) $\downarrow$ &  $l_2$ error  (b) $\downarrow$ \\ 
\midrule
{GenMM w/o DA w/o SAM} & $0.05$ & $0.0011$  && $1.08$ & $ 0.03$ \\
{GenMM w/o DA} & $0.048$ & $0.0003$  && $1.001$ & $ 0.0075$ \\
{GenMM}& $\textbf{0.025}$ & $\textbf{0.0002}$  && $\textbf{0.531}$ & $\textbf{0.0044}$ \\
\bottomrule
\end{tabular}
\caption{\lidar{} metrics for object insertion: columns (a) are computed for the \lidar{}  points on the inserted object, while columns (b) are computed across all the \lidar{} points in a frame.
AbsRel error computes mean absolute difference between the estimated and ground-truth range values, while the $l_2$ error computes the mean $l_2$ error between the estimated and the ground-truth 3D points.}
\label{tab:lidar_metrics}
\end{table*}

\begin{figure}[h]
\includegraphics[width=0.98\linewidth]{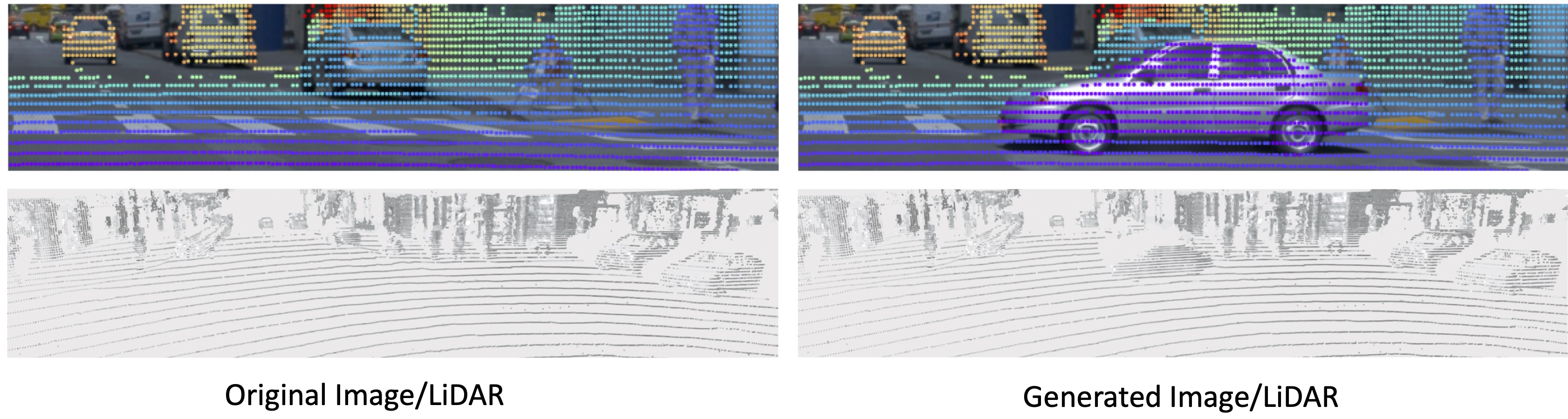}
\caption{We show zoomed in range (top row) and birds eye view (bottom row) for the \lidar{} points. Left column shows the original image/lidar where we insert a car. The top row highlights that our image and \lidar{} point clouds are aligned after synthesis. \lidar{} points with different color correspond to different depth, notice how \lidar{} color changes after car is introduced.}
\label{fig::lidar2}
\end{figure} 

\subsection{Evaluating Generated \lidar{} Points}
Figures~\ref{fig::lidar} and~\ref{fig::lidar2} show qualitative examples of \lidar{} outputs after inserting novel objects in the scene. As shown, the generated \lidar{} scenes are geometrically consistent and capture the shape of inserted object and visibility constraints (self-occlusions) faithfully. 

In order to quantitatively evaluate the quality of 3D lifted objects in the \lidar{} space, we take scenes with an existing object in 3D and remove it from the scene. The removed 3D object will serve as the ground truth. In Table~\ref{tab:lidar_metrics} we compare 3D reconstruction metrics of GenMM with two of its ablations, namely, GenMM w/o DA (i.e., GenMM without DepthAnything ~\cite{yang2024depth}), and GenMM w/o DA w/o SAM (i.e., GenMM without DepthAnything~\cite{yang2024depth} and SAM~\cite{kirillov2023segment}). 

We compare the reconstructed shape with each of these methods with the 3D ground truth using metrics such as AbsRel error = $\text{mean}_i [\|d_i^*-d_i\|/d_i$], which calculates the absolute relative depth error, and $l_2$ error = $\text{mean}_i\|\boldsymbol p_i^*-\boldsymbol p_i\|_2$ between updated and the ground-truth 3D \lidar{} points. Here, $d_i^*$ and $d_i$ are ground truth and reconstructed range values, respectively. Similarly, $\boldsymbol p_i^*$ and $\boldsymbol p_i$ are 3D ground truth and reconstructed points, respectively. As shown in the table, both DA and SAM enable better reconstruction (lower reconstruction errors) with the proposed GenMM method.

\section{Conclusion and Future Work}
We introduced GenMM, a method to generate multimodal data for object insertion in scenes using an object’s reference image rather than control primitives. Our approach is based on video-based diffusion models to ensure spatial and temporal coherence in the generated video sequences. Once images are generated, they serve as anchors for creating \lidar{} point clouds, which utilize monocular depth estimates derived from the images.

Future work will need to address the following limitations: currently, our model accounts only for geometry, not \lidar{} intensities, and assumes objects are opaque, which could reduce effectiveness for transparent surfaces like windows. 
As our work builds upon existing image generation methods, we inherit both the positive and negative societal impact of such methods. Our \lidar{} generation is conditioned on the inserted objects hence it cannot generate independent data and thus does not contribute any negative effects independently.


\begin{thebibliography}{10}

\bibitem{transfusion_2022}
Xuyang Bai, Zeyu Hu, Xinge Zhu, Qingqiu Huang, Yilun Chen, Hangbo Fu, and Chiew-Lan Tai.
\newblock Transfusion: Robust lidar-camera fusion for 3d object detection with transformers.
\newblock In {\em CVPR}, 2022.

\bibitem{arkitscenes}
Gilad Baruch, Zhuoyuan Chen, Afshin Dehghan, Tal Dimry, Yuri Feigin, Peter Fu, Thomas Gebauer, Brandon Joffe, Daniel Kurz, Arik Schwartz, and Elad Shulman.
\newblock Arkitscenes - a diverse real-world dataset for 3d indoor scene understanding using mobile rgb-d data.
\newblock In {\em NeuRIPS}, 2021.

\bibitem{blattmann2023stable}
Andreas Blattmann, Tim Dockhorn, Sumith Kulal, Daniel Mendelevitch, Maciej Kilian, Dominik Lorenz, Yam Levi, Zion English, Vikram Voleti, Adam Letts, et~al.
\newblock Stable video diffusion: Scaling latent video diffusion models to large datasets.
\newblock {\em CoRR}, abs/2311.15127, 2023.

\bibitem{videoworldsimulators2024}
Tim Brooks, Bill Peebles, Connor Holmes, Will DePue, Yufei Guo, Li~Jing, David Schnurr, Joe Taylor, Troy Luhman, Eric Luhman, Clarence Ng, Ricky Wang, and Aditya Ramesh.
\newblock Video generation models as world simulators.
\newblock 2024.

\bibitem{chen2023control}
Weifeng Chen, Jie Wu, Pan Xie, Hefeng Wu, Jiashi Li, Xin Xia, Xuefeng Xiao, and Liang Lin.
\newblock Control-a-video: Controllable text-to-video generation with diffusion models.
\newblock {\em CoRR}, abs/2305.13840, 2023.

\bibitem{Chen2016Multiview3O}
Xiaozhi Chen, Huimin Ma, Ji~Wan, Bo~Li, and Tian Xia.
\newblock Multi-view 3d object detection network for autonomous driving.
\newblock In {\em CVPR}, 2016.

\bibitem{chen2021geosim}
Yun Chen, Frieda Rong, Shivam Duggal, Shenlong Wang, Xinchen Yan, Sivabalan Manivasagam, Shangjie Xue, Ersin Yumer, and Raquel Urtasun.
\newblock Geosim: Realistic video simulation via geometry-aware composition for self-driving.
\newblock In {\em CVPR}, 2021.

\bibitem{carla_2017}
Alexey Dosovitskiy, German Ros, Felipe Codevilla, Antonio Lopez, and Vladlen Koltun.
\newblock {CARLA}: {An} open urban driving simulator.
\newblock In {\em Proceedings of the 1st Annual Conference on Robot Learning}, 2017.

\bibitem{unrealengine}
{Epic Games}.
\newblock Unreal engine, 2024.
\newblock Available online: \url{https://www.unrealengine.com/}.

\bibitem{ransac_1981}
M.~Fischler and R.~Bolles.
\newblock Random sample consensus: A paradigm for model fitting with applications to image analysis and automated cartography.
\newblock {\em Communications of the ACM}, 1981.

\bibitem{virtual_kitti}
Adrien Gaidon, Qiao Wang, Yohann Cabon, and Eleonora Vig.
\newblock Virtualworlds as proxy for multi-object tracking analysis.
\newblock {\em CVPR}, 2016.

\bibitem{geyer2023tokenflow}
Michal Geyer, Omer Bar-Tal, Shai Bagon, and Tali Dekel.
\newblock Tokenflow: Consistent diffusion features for consistent video editing.
\newblock {\em CoRR}, abs/2307.10373, 2023.

\bibitem{guo2023animatediff}
Yuwei Guo, Ceyuan Yang, Anyi Rao, Yaohui Wang, Yu~Qiao, Dahua Lin, and Bo~Dai.
\newblock Animatediff: Animate your personalized text-to-image diffusion models without specific tuning.
\newblock {\em CoRR}, abs/2307.04725, 2023.

\bibitem{ho2020denoising}
Jonathan Ho, Ajay Jain, and Pieter Abbeel.
\newblock Denoising diffusion probabilistic models.
\newblock In {\em NeuRIPS}, 2020.

\bibitem{hu2023gaia}
Anthony Hu, Lloyd Russell, Hudson Yeo, Zak Murez, George Fedoseev, Alex Kendall, Jamie Shotton, and Gianluca Corrado.
\newblock Gaia-1: A generative world model for autonomous driving.
\newblock {\em CoRR}, abs/2309.17080, 2023.

\bibitem{hu2021lora}
Edward~J Hu, Yelong Shen, Phillip Wallis, Zeyuan Allen-Zhu, Yuanzhi Li, Shean Wang, Lu~Wang, and Weizhu Chen.
\newblock Lora: Low-rank adaptation of large language models.
\newblock {\em CoRR}, abs/2106.09685, 2021.

\bibitem{hu2023animate}
Li~Hu, Xin Gao, Peng Zhang, Ke~Sun, Bang Zhang, and Liefeng Bo.
\newblock Animate anyone: Consistent and controllable image-to-video synthesis for character animation.
\newblock {\em CoRR}, abs/2311.17117, 2023.

\bibitem{hu2023_uniad}
Yihan Hu, Jiazhi Yang, Li~Chen, Keyu Li, Chonghao Sima, Xizhou Zhu, Siqi Chai, Senyao Du, Tianwei Lin, Wenhai Wang, Lewei Lu, Xiaosong Jia, Qiang Liu, Jifeng Dai, Yu~Qiao, and Hongyang Li.
\newblock Planning-oriented autonomous driving.
\newblock In {\em CVPR}, 2023.

\bibitem{kirillov2023segment}
Alexander Kirillov, Eric Mintun, Nikhila Ravi, Hanzi Mao, Chloe Rolland, Laura Gustafson, Tete Xiao, Spencer Whitehead, Alexander~C Berg, Wan-Yen Lo, et~al.
\newblock Segment anything.
\newblock In {\em ICCV}, 2023.

\bibitem{kutulakos2000theory}
Kiriakos~N Kutulakos and Steven~M Seitz.
\newblock A theory of shape by space carving.
\newblock {\em International journal of computer vision}, 38, 2000.

\bibitem{li2023lift3d}
Leheng Li, Qing Lian, Luozhou Wang, Ningning Ma, and Ying-Cong Chen.
\newblock Lift3d: Synthesize 3d training data by lifting 2d gan to 3d generative radiance field.
\newblock In {\em Proceedings of the IEEE/CVF Conference on Computer Vision and Pattern Recognition}, pages 332--341, 2023.

\bibitem{liu2023grounding}
Shilong Liu, Zhaoyang Zeng, Tianhe Ren, Feng Li, Hao Zhang, Jie Yang, Chunyuan Li, Jianwei Yang, Hang Su, Jun Zhu, et~al.
\newblock Grounding dino: Marrying dino with grounded pre-training for open-set object detection.
\newblock {\em CoRR}, abs/2303.05499, 2023.

\bibitem{peft}
Sourab Mangrulkar, Sylvain Gugger, Lysandre Debut, Younes Belkada, Sayak Paul, and Benjamin Bossan.
\newblock Peft: State-of-the-art parameter-efficient fine-tuning methods.
\newblock \url{https://github.com/huggingface/peft}, 2022.

\bibitem{Manivasagam2020LiDARsimRL}
Sivabalan Manivasagam, Shenlong Wang, K.~Wong, Wenyuan Zeng, Mikita Sazanovich, Shuhan Tan, Binh Yang, Wei-Chiu Ma, and Raquel Urtasun.
\newblock Lidarsim: Realistic lidar simulation by leveraging the real world.
\newblock In {\em CVPR}, 2020.

\bibitem{mildenhall2021nerf}
Ben Mildenhall, Pratul~P Srinivasan, Matthew Tancik, Jonathan~T Barron, Ravi Ramamoorthi, and Ren Ng.
\newblock Nerf: Representing scenes as neural radiance fields for view synthesis.
\newblock {\em Communications of the ACM}, 65(1), 2021.

\bibitem{omniverse}
{NVIDIA Corporation}.
\newblock Nvidia omniverse, 2024.
\newblock Available online: \url{https://www.nvidia.com/en-us/omniverse/}.

\bibitem{poole2022dreamfusion}
Ben Poole, Ajay Jain, Jonathan~T Barron, and Ben Mildenhall.
\newblock Dreamfusion: Text-to-3d using 2d diffusion.
\newblock {\em CoRR}, abs/2209.14988, 2022.

\bibitem{RadfordCLIP}
Alec Radford, Jong~Wook Kim, Chris Hallacy, Aditya Ramesh, Gabriel Goh, Sandhini Agarwal, Girish Sastry, Amanda Askell, Pamela Mishkin, Jack Clark, Gretchen Krueger, and Ilya Sutskever.
\newblock Learning transferable visual models from natural language supervision.
\newblock {\em CoRR}, abs/2103.00020, 2021.

\bibitem{Roberts_2021_ICCV}
Mike Roberts, Jason Ramapuram, Anurag Ranjan, Atulit Kumar, Miguel~Angel Bautista, Nathan Paczan, Russ Webb, and Joshua~M. Susskind.
\newblock Hypersim: A photorealistic synthetic dataset for holistic indoor scene understanding.
\newblock In {\em ICCV}, 2021.

\bibitem{rombach2022high}
Robin Rombach, Andreas Blattmann, Dominik Lorenz, Patrick Esser, and Bj{\"o}rn Ommer.
\newblock High-resolution image synthesis with latent diffusion models.
\newblock In {\em CVPR}, 2022.

\bibitem{ruiz2023dreambooth}
Nataniel Ruiz, Yuanzhen Li, Varun Jampani, Yael Pritch, Michael Rubinstein, and Kfir Aberman.
\newblock Dreambooth: Fine tuning text-to-image diffusion models for subject-driven generation.
\newblock In {\em CVPR}, 2023.

\bibitem{sara2019image}
Umme Sara, Morium Akter, and Mohammad~Shorif Uddin.
\newblock Image quality assessment through fsim, ssim, mse and psnr—a comparative study.
\newblock {\em Journal of Computer and Communications}, 7(3), 2019.

\bibitem{savva2019habitat}
Manolis Savva, Abhishek Kadian, Oleksandr Maksymets, Yili Zhao, Erik Wijmans, Bhavana Jain, Julian Straub, Jia Liu, Vladlen Koltun, Jitendra Malik, et~al.
\newblock Habitat: A platform for embodied ai research.
\newblock In {\em ICCV}, 2019.

\bibitem{schulz2018interaction}
Jens Schulz, Constantin Hubmann, Julian L{\"o}chner, and Darius Burschka.
\newblock Interaction-aware probabilistic behavior prediction in urban environments.
\newblock In {\em IROS}, 2018.

\bibitem{shen2023gina}
Bokui Shen, Xinchen Yan, Charles~R Qi, Mahyar Najibi, Boyang Deng, Leonidas Guibas, Yin Zhou, and Dragomir Anguelov.
\newblock Gina-3d: Learning to generate implicit neural assets in the wild.
\newblock In {\em Proceedings of the IEEE/CVF conference on computer vision and pattern recognition}, pages 4913--4926, 2023.

\bibitem{simgan_2017_CVPR}
Ashish Shrivastava, Tomas Pfister, Oncel Tuzel, Joshua Susskind, Wenda Wang, and Russell Webb.
\newblock Learning from simulated and unsupervised images through adversarial training.
\newblock In {\em CVPR}, 2017.

\bibitem{singh2018sniper}
Bharat Singh, Mahyar Najibi, and Larry~S Davis.
\newblock Sniper: Efficient multi-scale training.
\newblock {\em NeuRIPS}, 2018.

\bibitem{sun2020scalability}
Pei Sun, Henrik Kretzschmar, Xerxes Dotiwalla, Aurelien Chouard, Vijaysai Patnaik, Paul Tsui, James Guo, Yin Zhou, Yuning Chai, Benjamin Caine, et~al.
\newblock Scalability in perception for autonomous driving: Waymo open dataset.
\newblock In {\em CVPR}, 2020.

\bibitem{suo2021trafficsim}
Simon Suo, Sebastian Regalado, Sergio Casas, and Raquel Urtasun.
\newblock Trafficsim: Learning to simulate realistic multi-agent behaviors.
\newblock In {\em CVPR}, 2021.

\bibitem{tang2023make}
Junshu Tang, Tengfei Wang, Bo~Zhang, Ting Zhang, Ran Yi, Lizhuang Ma, and Dong Chen.
\newblock Make-it-3d: High-fidelity 3d creation from a single image with diffusion prior.
\newblock In {\em ICCV}, 2023.

\bibitem{unity3d}
{Unity Technologies}.
\newblock Unity, 2024.
\newblock Available online: \url{https://unity.com/}.

\bibitem{unterthiner2019fvd}
Thomas Unterthiner, Sjoerd van Steenkiste, Karol Kurach, Rapha{\"e}l Marinier, Marcin Michalski, and Sylvain Gelly.
\newblock {FVD}: A new metric for video generation, 2019.

\bibitem{point_painting_2020}
Sourabh Vora, Alex~H. Lang, Bassam Helou, and Oscar Beijbom.
\newblock Pointpainting: Sequential fusion for 3d object detection.
\newblock In {\em CVPR}, 2020.

\bibitem{wang2023score}
Haochen Wang, Xiaodan Du, Jiahao Li, Raymond~A Yeh, and Greg Shakhnarovich.
\newblock Score jacobian chaining: Lifting pretrained 2d diffusion models for 3d generation.
\newblock In {\em CVPR}, 2023.

\bibitem{wu2023mars}
Zirui Wu, Tianyu Liu, Liyi Luo, Zhide Zhong, Jianteng Chen, Hongmin Xiao, Chao Hou, Haozhe Lou, Yuantao Chen, Runyi Yang, et~al.
\newblock Mars: An instance-aware, modular and realistic simulator for autonomous driving.
\newblock In {\em CAAI International Conference on Artificial Intelligence}, 2023.

\bibitem{yang2023emernerf}
Jiawei Yang, Boris Ivanovic, Or~Litany, Xinshuo Weng, Seung~Wook Kim, Boyi Li, Tong Che, Danfei Xu, Sanja Fidler, Marco Pavone, et~al.
\newblock Emernerf: Emergent spatial-temporal scene decomposition via self-supervision.
\newblock {\em CoRR}, abs/2311.02077, 2023.

\bibitem{yang2024depth}
Lihe Yang, Bingyi Kang, Zilong Huang, Xiaogang Xu, Jiashi Feng, and Hengshuang Zhao.
\newblock Depth anything: Unleashing the power of large-scale unlabeled data.
\newblock {\em CoRR}, abs/2401.10891, 2024.

\bibitem{yang2023unisim}
Ze~Yang, Yun Chen, Jingkang Wang, Sivabalan Manivasagam, Wei-Chiu Ma, Anqi~Joyce Yang, and Raquel Urtasun.
\newblock Unisim: A neural closed-loop sensor simulator.
\newblock In {\em CVPR}, pages 1389--1399, 2023.

\bibitem{yin2024isfusion}
Junbo Yin, Jianbing Shen, Runnan Chen, Wei Li, Ruigang Yang, Pascal Frossard, and Wenguan Wang.
\newblock {IS-FUSION}: Instance-scene collaborative fusion for multimodal 3d object detection.
\newblock In {\em CVPR}, 2024.

\bibitem{yu2020bdd100k}
Fisher Yu, Haofeng Chen, Xin Wang, Wenqi Xian, Yingying Chen, Fangchen Liu, Vashisht Madhavan, and Trevor Darrell.
\newblock Bdd100k: A diverse driving dataset for heterogeneous multitask learning.
\newblock In {\em CVPR}, 2020.

\bibitem{controlnet_iccv_2023}
Lvmin Zhang, Anyi Rao, and Maneesh Agrawala.
\newblock Adding conditional control to text-to-image diffusion models.
\newblock In {\em ICCV}, 2023.

\bibitem{zhang2018perceptual}
Richard Zhang, Phillip Isola, Alexei~A Efros, Eli Shechtman, and Oliver Wang.
\newblock The unreasonable effectiveness of deep features as a perceptual metric.
\newblock In {\em CVPR}, 2018.

\bibitem{zhou2023drivinggaussian}
Xiaoyu Zhou, Zhiwei Lin, Xiaojun Shan, Yongtao Wang, Deqing Sun, and Ming-Hsuan Yang.
\newblock Drivinggaussian: Composite gaussian splatting for surrounding dynamic autonomous driving scenes.
\newblock {\em CoRR}, abs/2312.07920, 2023.

\end{thebibliography}

\clearpage
\appendix

\section{Appendix}
Here, we show more visual examples for object insertion, swapping, animation and multi-modal data generation. 

\begin{figure}[h]
\includegraphics[width=\linewidth]{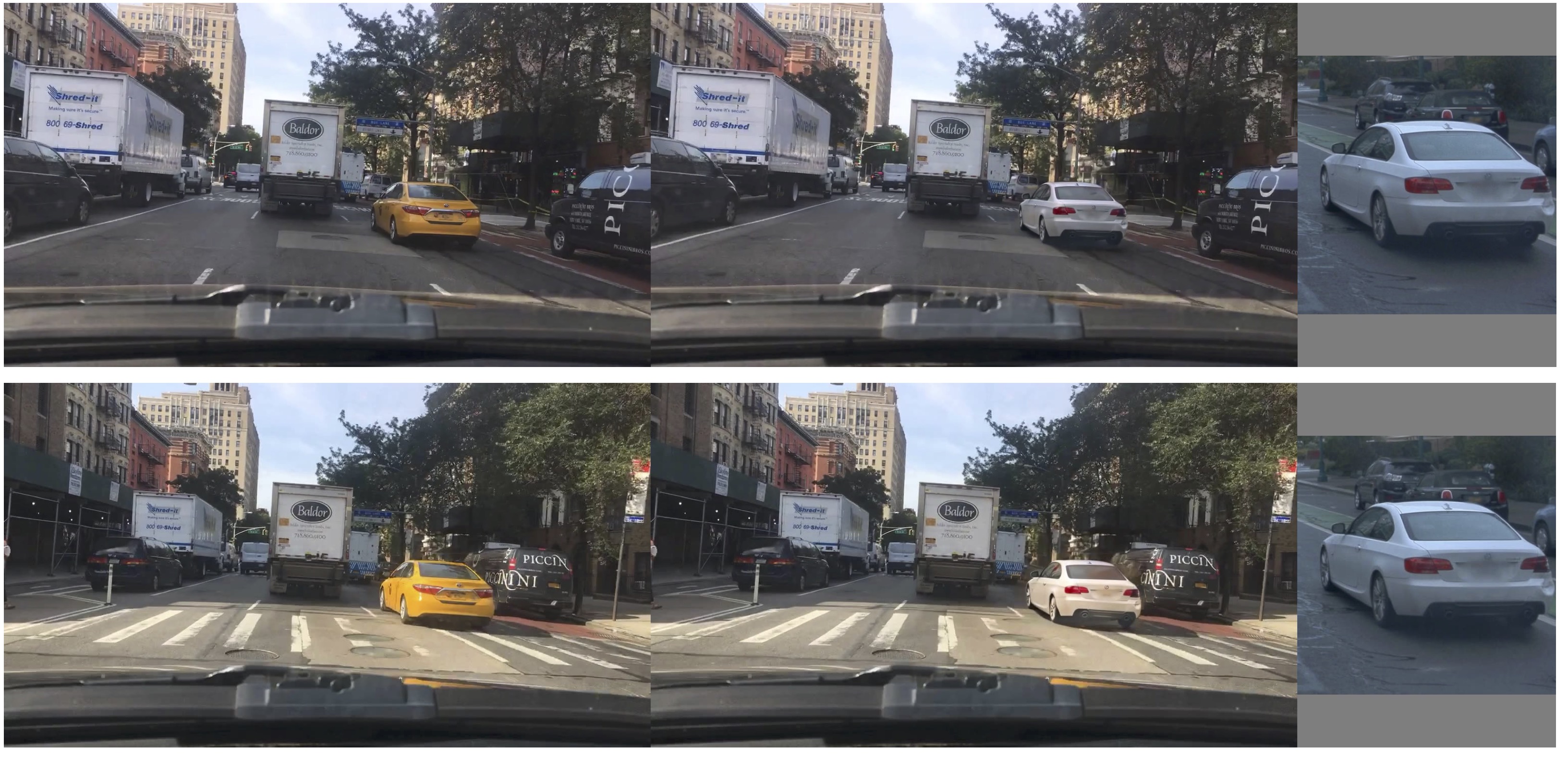}
\caption{Video object swapping. Left: input, Middle: generated, Right: reference objects. A yellow car replaced by a white car. 
Note the realistic rendering relighting of the newly added car in sun and shadows, despite the reference image having different lighting conditions.}
\label{fig::swapping_sun_and_shadows}
\end{figure} 

\begin{figure}[h]
\includegraphics[width=\linewidth]{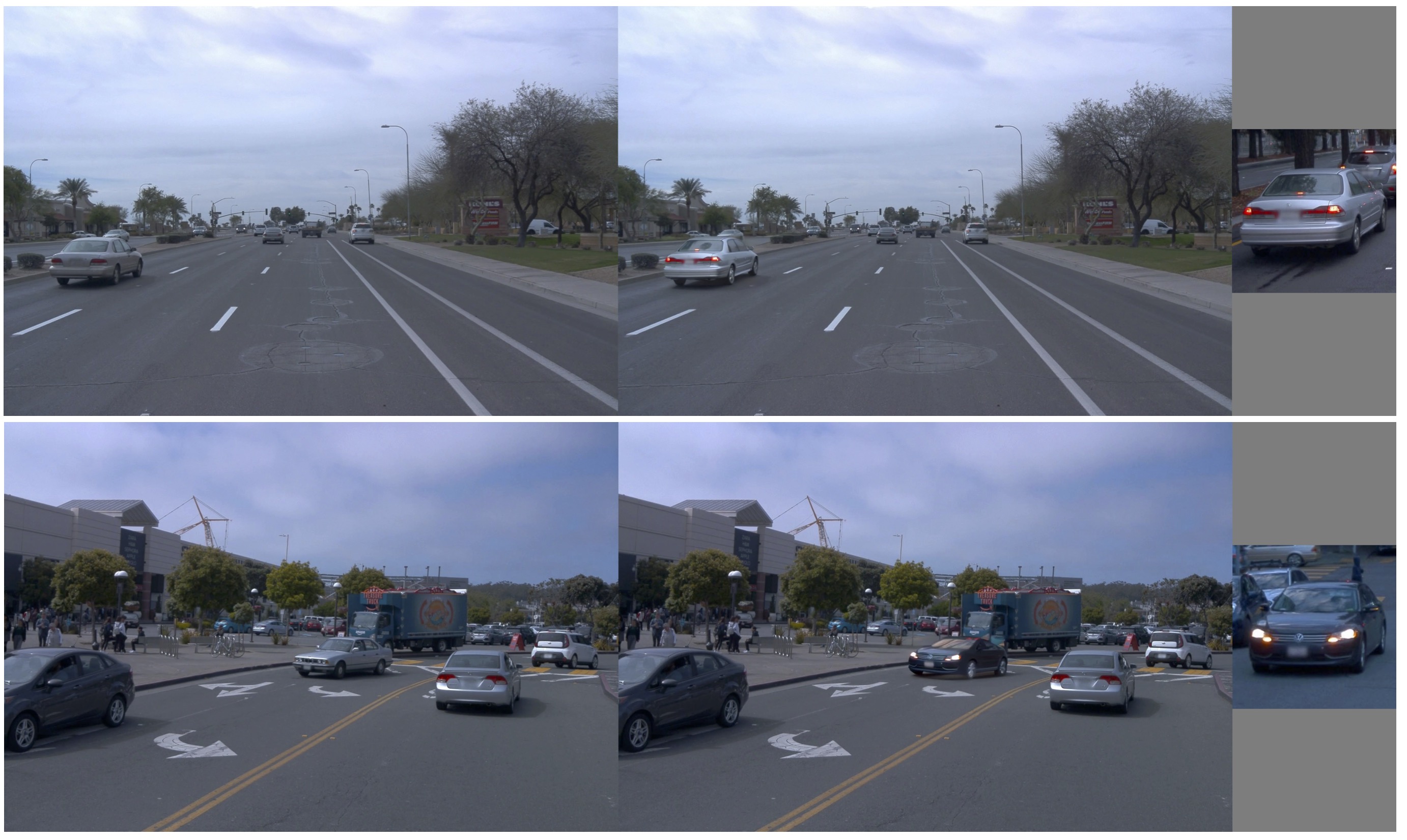}
\caption{Video object swapping. Left: input frames, Middle: generated frames, Right: reference objects. A car with tail/headlights off being replaced by another car with tail/headlights on.}
\label{fig::swapping_llights_on}
\end{figure}

\begin{figure}[h]
\includegraphics[width=\linewidth]{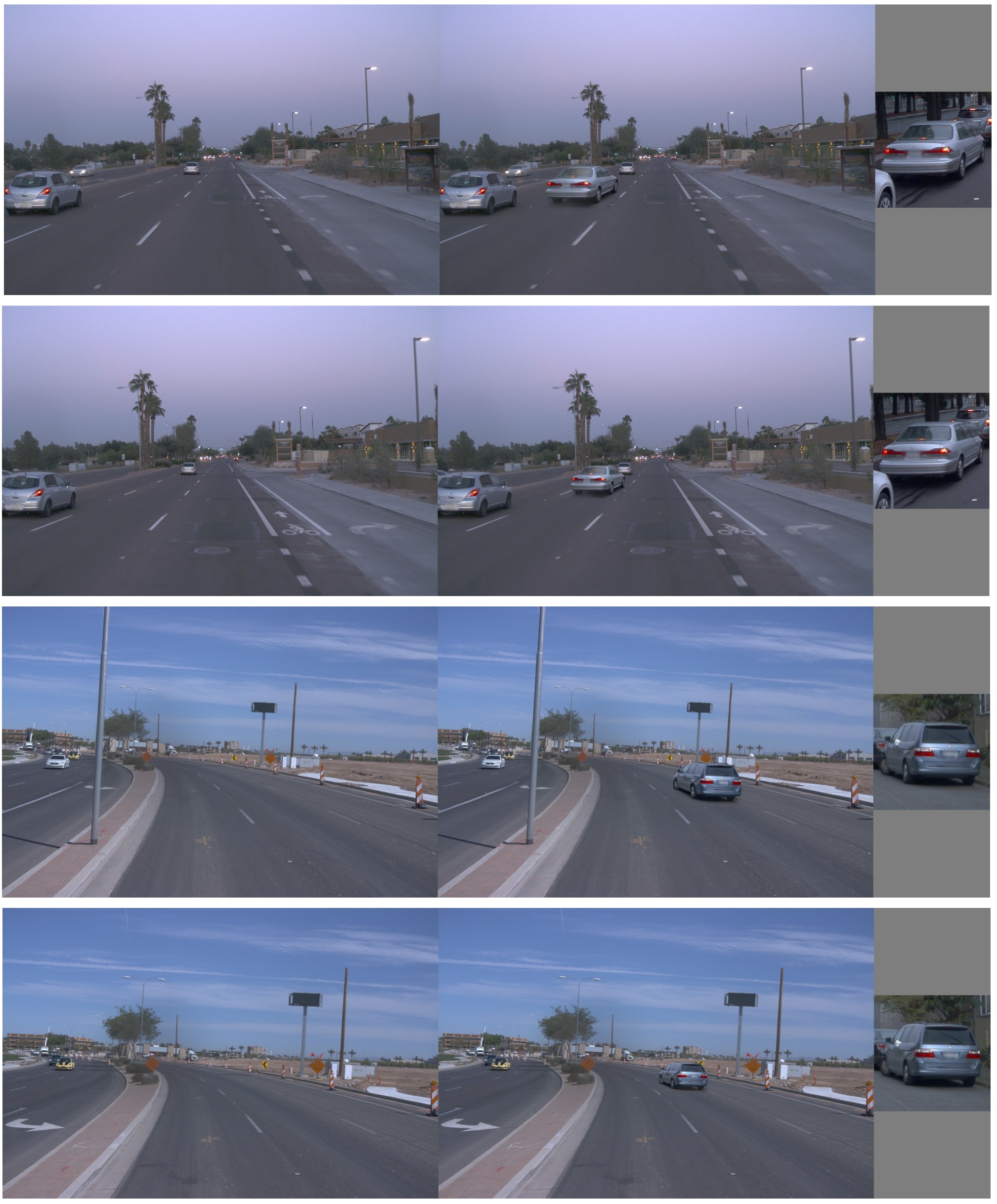}
\caption{Video object insertion. Left: input frames, Middle: generated frames, Right: reference objects. We insert a car and a minivan in a video and show two different frames for this task.}
\label{fig::object_insertion_1}
\end{figure} 

\begin{figure}[h]
\includegraphics[width=\linewidth]{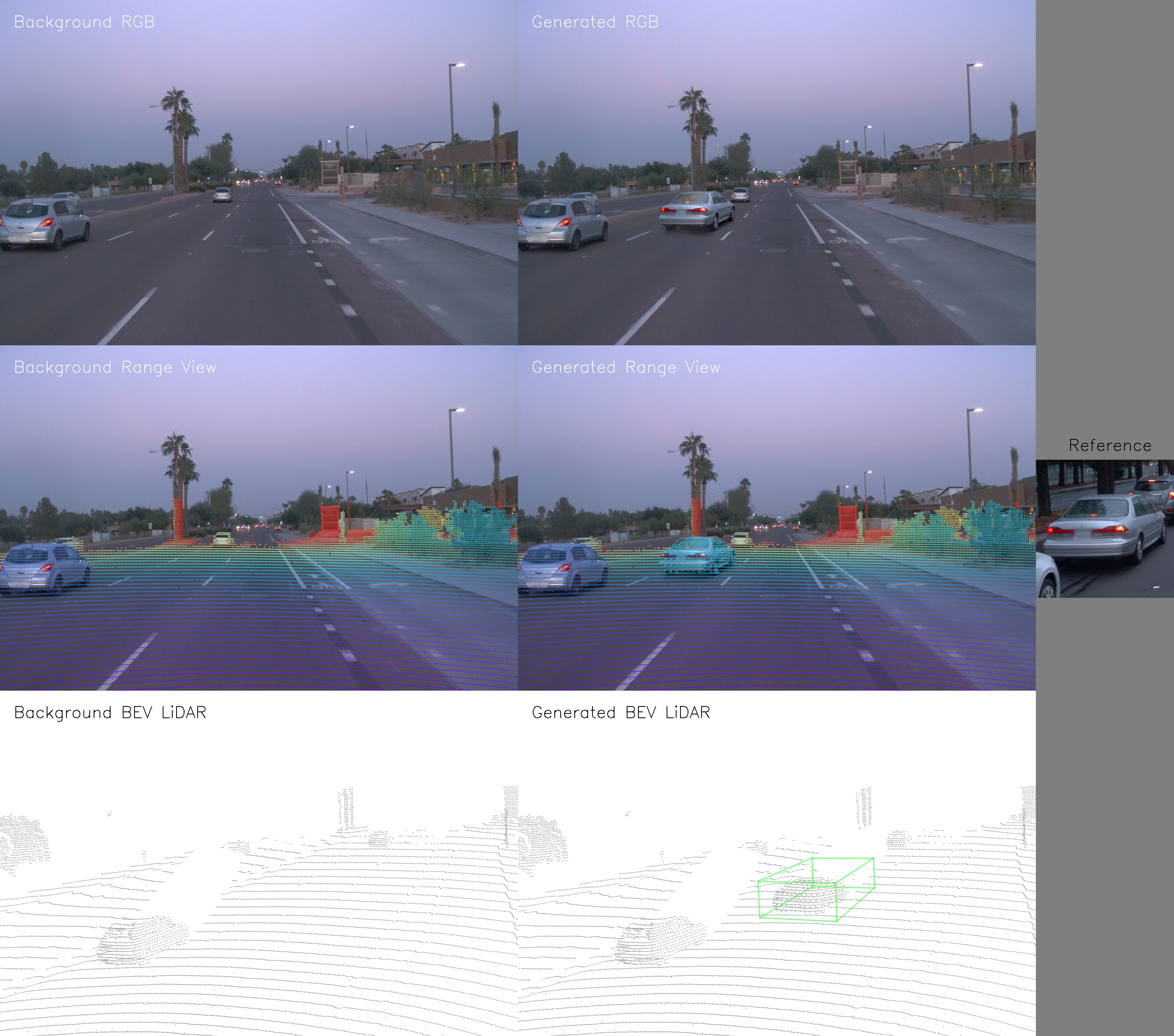}
\caption{Multi-modal Video object insertion. Left: input frame, Middle: generated frame, Right: reference car. We insert a back view of a car and show its corresponding LiDAR map. The middle row highlights that our image and \lidar{} point clouds are aligned after synthesis.}
\label{fig::object_insertion_2}
\end{figure} 

\begin{figure}[h]
\includegraphics[width=\linewidth]{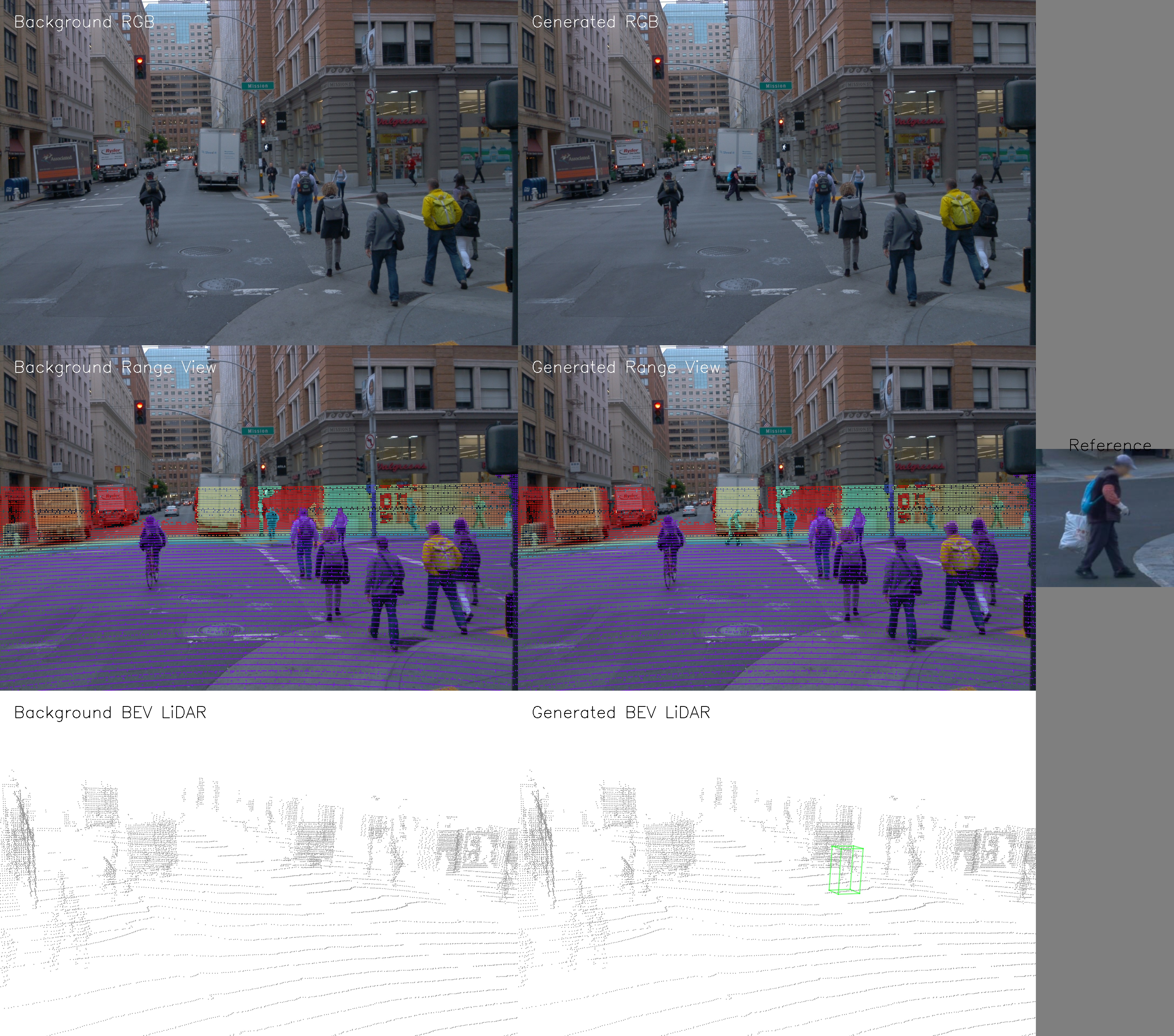}
\caption{Multi-modal Video object insertion. Left: input frame, Middle: generated frame, Right: reference person. We insert a person in front of the white truck and show its corresponding LiDAR map. The middle row highlights that our image and \lidar{} point clouds are aligned after synthesis, also notice the change in color of the lidar pixels on person to green as the lidar range changes for the inserted person.}
\label{fig::object_insertion_3}
\end{figure} 

\begin{figure}[h]
\includegraphics[width=\linewidth]{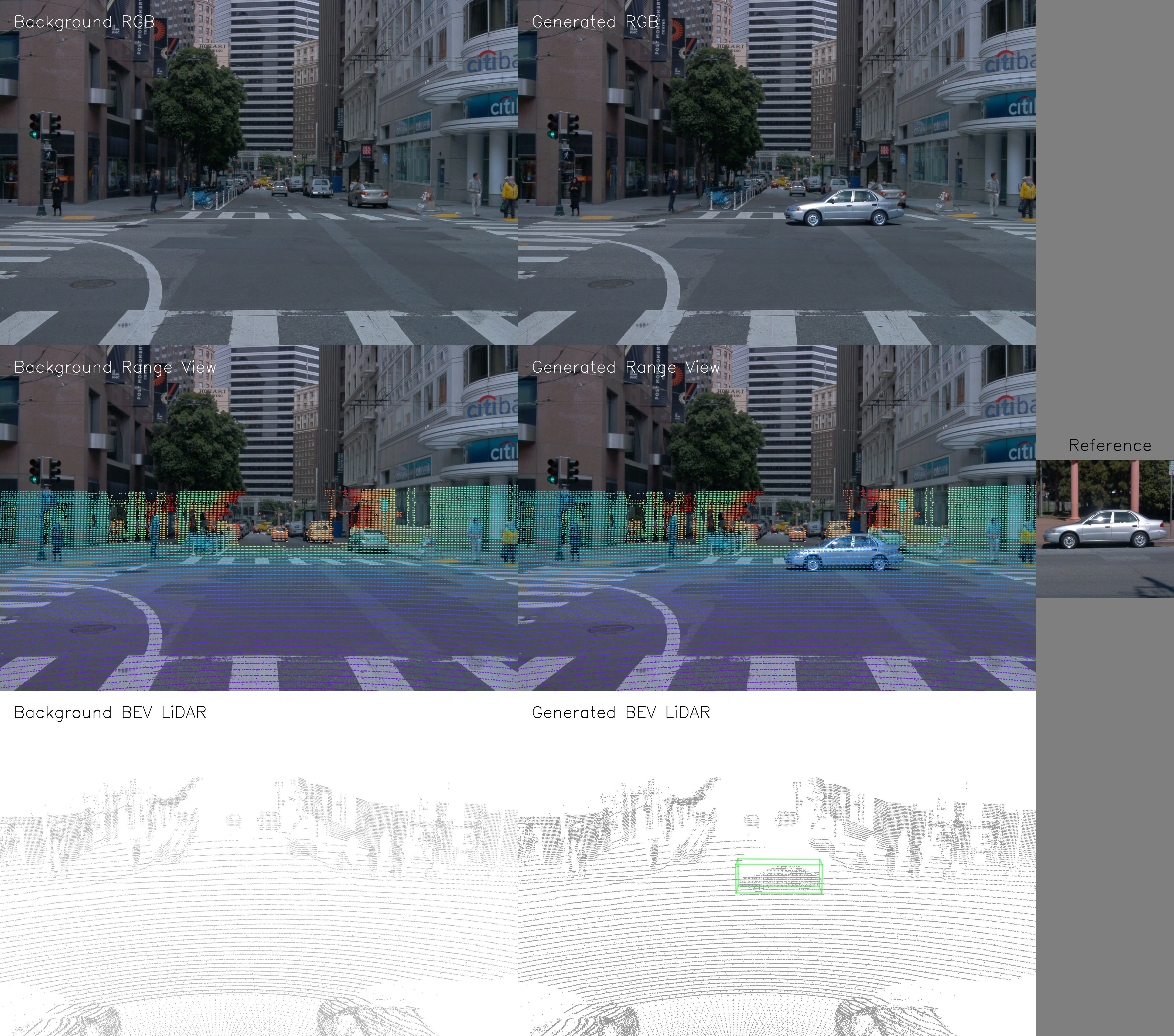}
\caption{Multi-modal Video object insertion. Left: input frame, Middle: generated frame, Right: reference car. We insert a side view of a car and show its corresponding LiDAR map. The middle row highlights that our image and \lidar{} point clouds are aligned after synthesis.}
\label{fig::object_insertion_4}
\end{figure}

\begin{figure}[h]
\includegraphics[width=\linewidth]{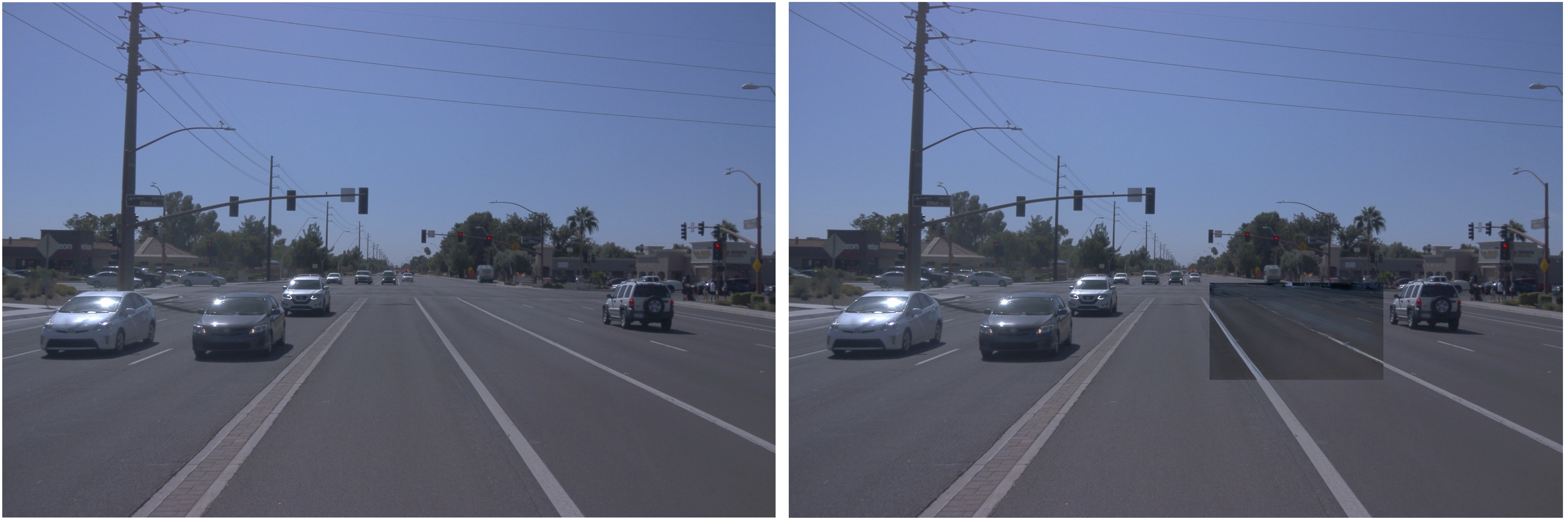}
\caption{Visual results for object insertion using Tokenflow. As we can see, Tokenflow ends up generating background pixels.}
\label{fig::tokenflow}
\end{figure}

\end{document}